\documentclass[conference]{IEEEtran}
\IEEEoverridecommandlockouts

 \usepackage{units}
 \usepackage{bm}
 \usepackage{float}
 \usepackage{booktabs}
 \usepackage{graphicx}
 \usepackage{caption}
 \usepackage{subcaption}
 \usepackage{amsmath} 
 \usepackage{amssymb}  
 \usepackage{mathtools}
 \graphicspath{ {figures/} }
 \usepackage[acronym,nomain,nonumberlist]{glossaries}

 \newacronym{mav}{MAV}{Micro Aerial Vehicles}
\newacronym{uav}{UAV}{Unmanned Aerial Vehicle}
\newacronym{nmpc}{NMPC}{Nonlinear Model Predictive Control}
\newacronym{mpc}{MPC}{Model Predictive Control}
\newacronym{panoc}{PANOC}{Proximal Averaged Newton-type method for Optimal Control}
\newacronym{lidar}{LiDAR}{Light Detection And Ranging}
\newacronym{pcl}{PCL}{Point Cloud}
\newacronym{sub-t}{Sub-T}{Sub-Terranean}
\newacronym{gnss}{GNSS}{Global Navigation Satellite System}
\newacronym{slam}{SLAM}{Simultaneous Localization and Mapping}
\newacronym{pde}{PDE}{Perceptually Degraded Environments}
\newacronym{hri}{HRI}{Human-robot Interaction}
\newacronym{hsi}{HSI}{Human-swarm Interaction}
\newacronym{os}{OS}{Operating System}
\newacronym{fov}{FOV}{Field of View}
\newacronym{imu}{IMU}{Inertial Measurement Unit}
\newacronym{ros}{ROS}{Robotic Operating System}
\newacronym{sor}{SOR}{Statistical Outlier Removal}
\newacronym{ror}{ROR}{Radius Outlier Removal}
\newacronym{dsor}{DSOR}{Dynamic Statistical Outlier Removal}
\newacronym{dror}{DROR}{Dynamic Radius Outlier Removal}
\newacronym{lior}{LiOR}{Low-intensity Outlier Removal}
\newacronym{lidror}{LiDROR}{Low-Intensity Dynamic Radius Outlier Removal}
\newacronym{ml}{ML}{Machine Learning}
\newacronym{knn}{k-NN}{k-Nearest Neighbor}
\newacronym{svm}{SVM}{Support Vector Machine}
\newacronym{rf}{RF}{Random Forest}
\newacronym{dbscan}{DBSCAN}{Density-based Spatial Clustering of Applications with Noise}
\newacronym{nn}{NN}{Neural Network}
\newacronym{cnn}{CNN}{Convolutional Neural Network}
\newacronym{rss}{RSS}{Responsibility-sensitive Safety}
\newacronym{sg}{SG}{Savitzky–Golay}
\newacronym{sota}{SOTA}{state-of-the-art}
\newacronym{sar}{SAR}{Search and Rescue}
\newacronym{ldmm}{LDMM}{Low-Dimensional Manifold Model}
\newacronym{mls}{MLS}{Moving Least Squares} 
\newacronym{apf}{APF}{Artificial Potential Field}
\newacronym{ccs}{CCS}{Cartesian Coordinate System}
\newacronym{scs}{SCS}{Spherical Coordinate System}
\newacronym{ccf}{CCF}{Cartesian Coordinate Frame}
\newacronym{doscor}{DoSCOR}{Dynamic onboard Statistical Cluster Outlier Removal}
\newacronym{pdf}{PDF}{Probability Density Function}
\newacronym{soa}{SoA}{State-of-the-Art}
\newacronym{dlo}{DLO}{Direct LiDAR Odometry}
\newacronym{liosam}{LIO-SAM}{LiDAR Inertial Odometry via Smoothing and Mapping}
\newacronym{icp}{ICP}{Iterative Closest Point}
\newacronym{qf}{QF}{Quantile Function}

\usepackage{cite}
\usepackage{amsmath,amssymb,amsfonts}
\usepackage{algorithmic}
\usepackage{graphicx}
\usepackage{textcomp}
\usepackage{xcolor}
\def\BibTeX{{\rm B\kern-.05em{\sc i\kern-.025em b}\kern-.08em
    T\kern-.1667em\lower.7ex\hbox{E}\kern-.125emX}}
\begin{document}

\title{Efficient Real-time Smoke Filtration with 3D LiDAR for Search and Rescue with Autonomous Heterogeneous Robotic Systems\\
\thanks{This work has been partially funded by the European Unions Horizon 2020 Research and Innovation Programme under the Grant Agreement No. 101003591 NEX-GEN SIMS. Corresponding author email:{\tt\small akyuroson@gmail.com}}
}

\author{\IEEEauthorblockN{Alexander Kyuroson\IEEEauthorrefmark{1}, Anton Koval\IEEEauthorrefmark{1} and George Nikolakopoulos\IEEEauthorrefmark{1}}
\IEEEauthorblockA{\IEEEauthorrefmark{1}Robotics and Artificial Intelligence Group,\\ Department of Computer, Electrical and Space Engineering,\\ Lule\r{a} University of Technology, Lule\r{a} SE-97187, Sweden\\
Email: akyuroson@gmail.com}
}


\maketitle

\begin{abstract}
\gls{sar} missions in harsh and unstructured~\gls{sub-t} environments in the presence of aerosol particles have recently become the main focus in the field of robotics. Aerosol particles such as smoke and dust directly affect the performance of any mobile robotic platform due to their reliance on their onboard perception systems for autonomous navigation and localization in~\gls{gnss}-denied environments. Although obstacle avoidance and object detection algorithms are robust to the presence of noise to some degree, their performance directly relies on the quality of captured data by onboard sensors such as~\gls{lidar} and camera. Thus, this paper proposes a novel modular agnostic filtration pipeline based on intensity and spatial information such as local point density for removal of detected smoke particles from~\gls{pcl} prior to its utilization for collision detection. Furthermore, the efficacy of the proposed framework in the presence of smoke during multiple frontier exploration missions is investigated while the experimental results are presented to facilitate comparison with other methodologies and their computational impact. This provides valuable insight to the research community for better utilization of filtration schemes based on available computation resources while considering the safe autonomous navigation of mobile robots.
\end{abstract}

\begin{IEEEkeywords}
outlier rejection, aerosol particles, heterogeneous robotic systems
\end{IEEEkeywords}

\section{Introduction and Background} \label{introduction}
In recent years, hybrid-robotic systems with multi-sensor payloads have been deployed in harsh~\gls{sar} scenarios~\cite{lindqvist2022multimodality} to not only aid in the exploration of the~\gls{pde}s~\cite{mansouri2020deploying} by inspecting the environmental and structural conditions~\cite{dharmadhikari2021autonomous} but also assist rescue workers by increasing their situational awareness to improve rescue efforts while ensuring their safety~\cite{rafal2022sar} in such time-critical operations. 

\begin{figure}[!hb]
    \centering
    \includegraphics[width=0.69\columnwidth]{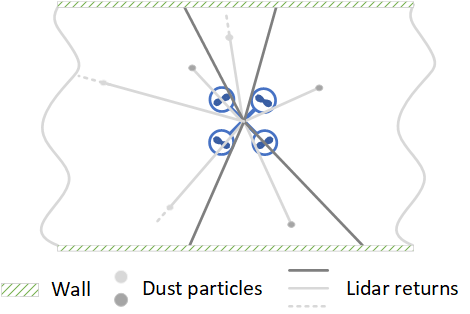}
    \setlength{\abovecaptionskip}{-0.25pt}
    \setlength{\belowcaptionskip}{-8pt}
    \caption{Behavior of~\gls{lidar} in the presence of dust, where the propulsion system of a drone entrains dust particles.}
    \label{fig:dust-drone-concept}
\end{figure}

Furthermore, such autonomous robotic platforms mainly rely on their onboard perception systems in~\gls{gnss}-denied environments such as~\gls{sub-t} and extra-terrestrial sub-surfaces~\cite{titus2021roadmap} for~\gls{slam}. To ensure operational safety in such hazardous environments, a combination of~\gls{lidar} and vision-based sensors is utilized to perform pose estimation~\cite{Zhao2021} and collision avoidance~\cite{lindqvist2020nonlinear}. The presence of aerosol particles such as smoke~\cite{lu2020see} and dust~\cite{Zhao2021} directly affects the performance of these sensors and increases the noise in captured data. In particular,~\gls{lidar} sensors are negatively influenced by aerosol particles which cause undesired measurements~\cite{phillips2017dust} of the laser beam thereby resulting in occlusions and additional cluttered and noisy points in generated~\gls{pcl}s~\cite{fritsche2018fusion}; see Figure~\ref{fig:dust-drone-concept}. 

It must be noted that~\gls{lidar}s are more prone to such issues in comparison to RADARs~\cite{khader2020introduction} due to their inherent beam divergence and short pulse duration~\cite{phillips2017dust}. However, due to their performance in poorly-illuminated environments, accurate range measurements and higher spatial resolution~\cite{zhang2017low} compared to RADARs and RGB-D cameras,~\gls{lidar}s have been deployed on many autonomous robotic platforms~\cite{lindqvist2021compra}.

The acquired~\gls{pcl}s are not only used for autonomous navigation of complex environments~\cite{lindqvist2021compra} but also to monitor environmental changes~\cite{lei2022tunnel}, detect artifacts~\cite{patel2022object}, and assess traversability~\cite{zhao2022terrain}. Therefore, it is vital to identify and remove the points caused by aerosol particles prior to the utilization of data for any other downstream algorithms. To address the aforementioned issues, various~\gls{pcl} filtration methods have been proposed~\cite{zhou2022fullreview}. These methods can be mainly divided into either classical or learning-based approaches~\cite{qu2023mahalanobis}.


\begin{figure*}
     \centering
     \begin{subfigure}[b]{0.32\textwidth}
        \centering
        \includegraphics[width=\columnwidth, trim={8cm 11.2cm 4cm 5cm}, clip]{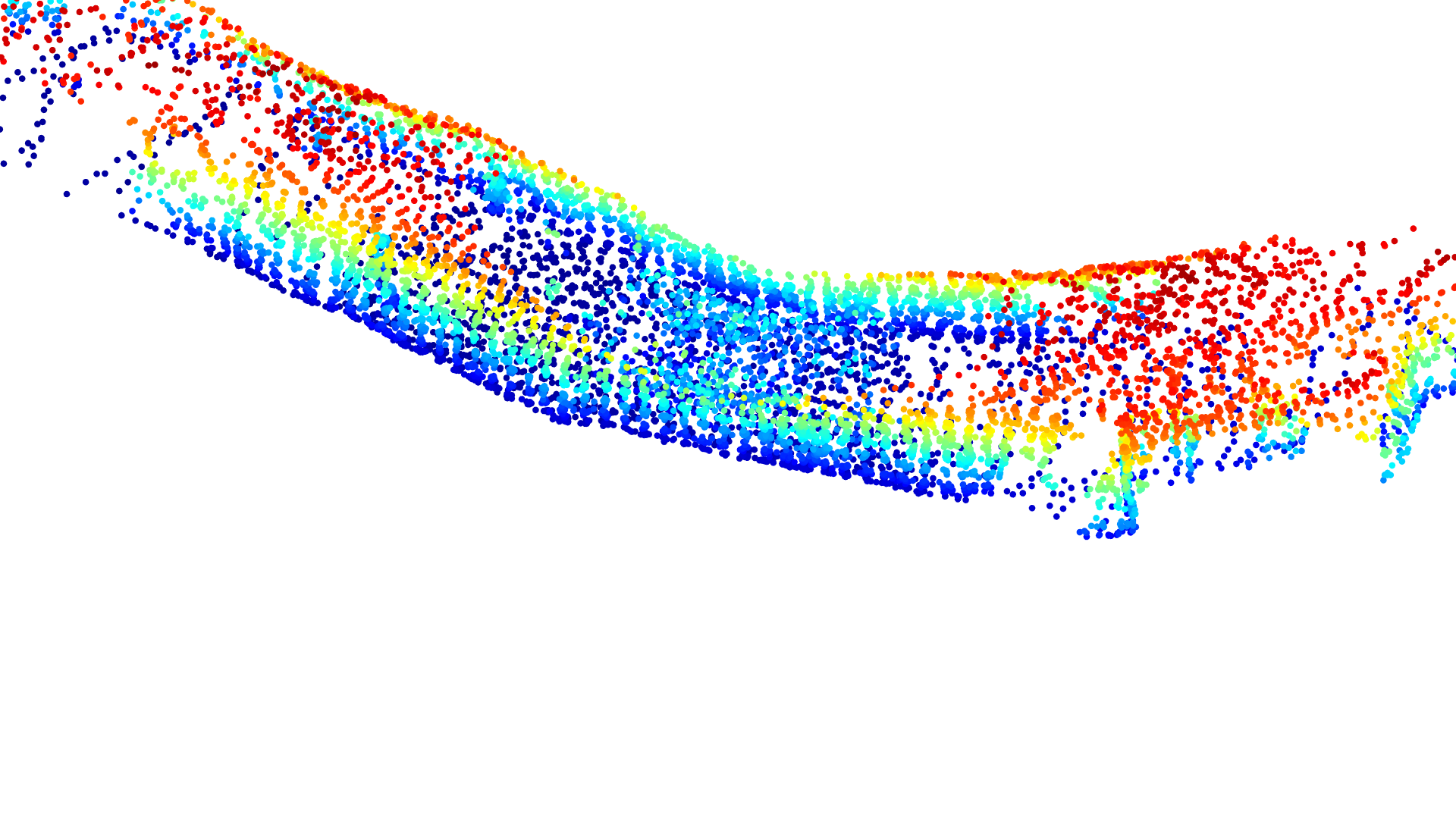}
        \caption{Raw~\gls{pcl} with dust.}
        \label{fig:fig1_raw_pcl}
     \end{subfigure}
     \hfill
    \begin{subfigure}[b]{0.32\textwidth}
        \centering
        \includegraphics[width=\columnwidth, trim={6cm 10cm 4cm 5cm}, clip]{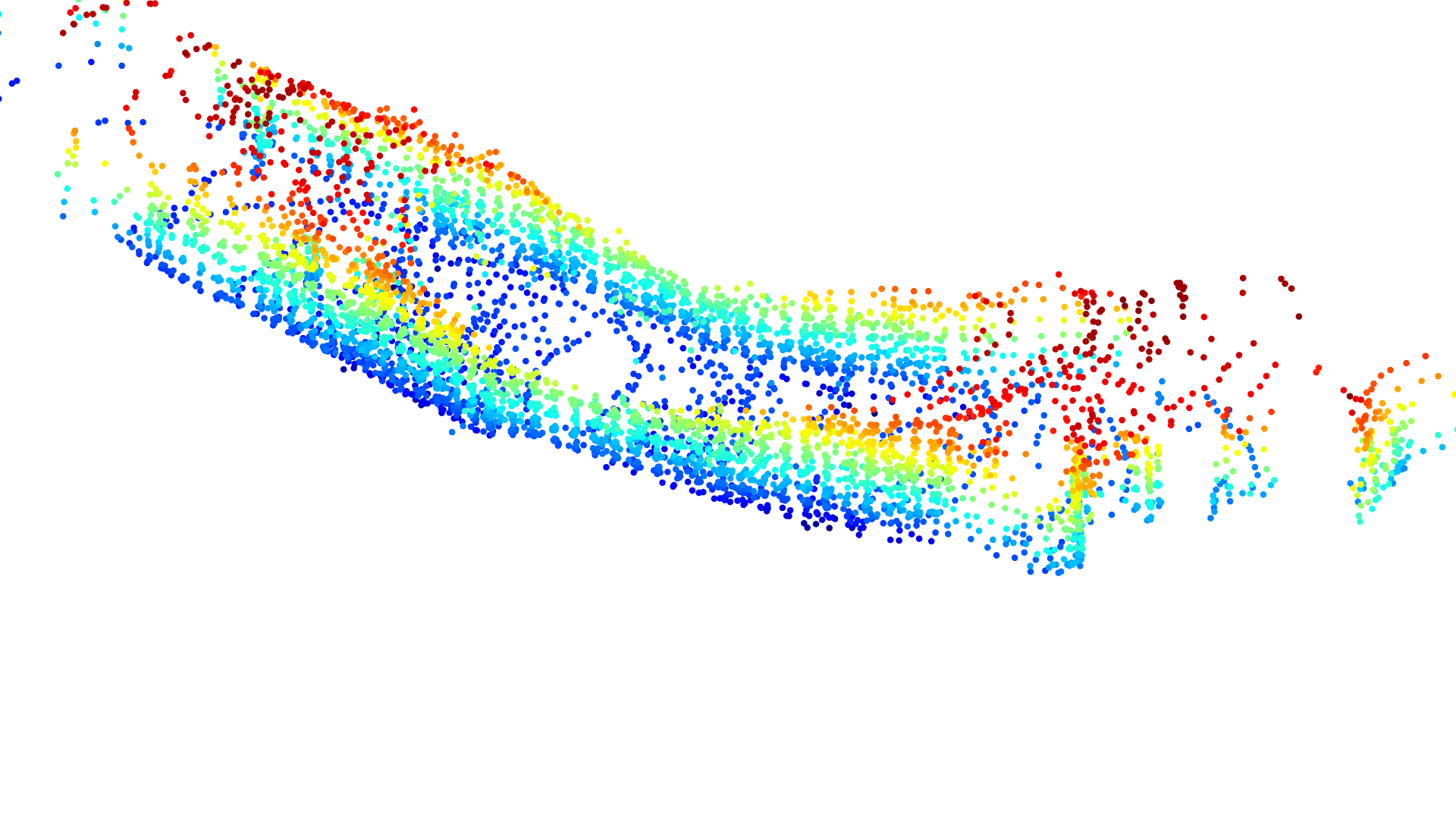}
        \caption{Our framework on.}
        \label{fig:fig1_filtered}
    \end{subfigure}
     \begin{subfigure}[b]{0.33\textwidth}
        \centering
        \includegraphics[width=\columnwidth, trim={7cm 8cm 12cm 7cm}, clip]{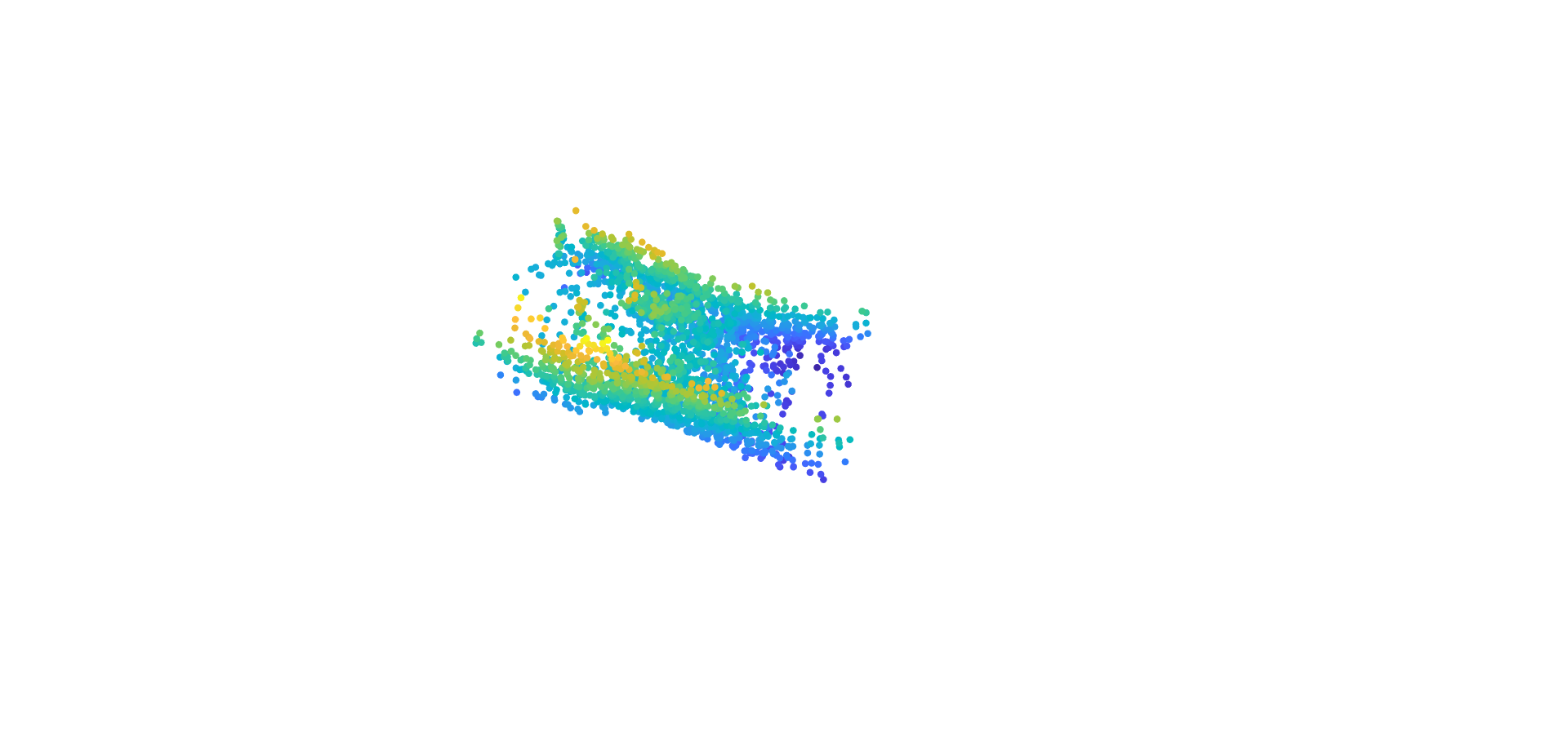}
        \caption{Dust isolated in the~\gls{pcl}.}
        \label{fig:fig1_dust}
     \end{subfigure}
    \begin{subfigure}[b]{0.315\textwidth}
        \centering
        \includegraphics[width=\columnwidth]{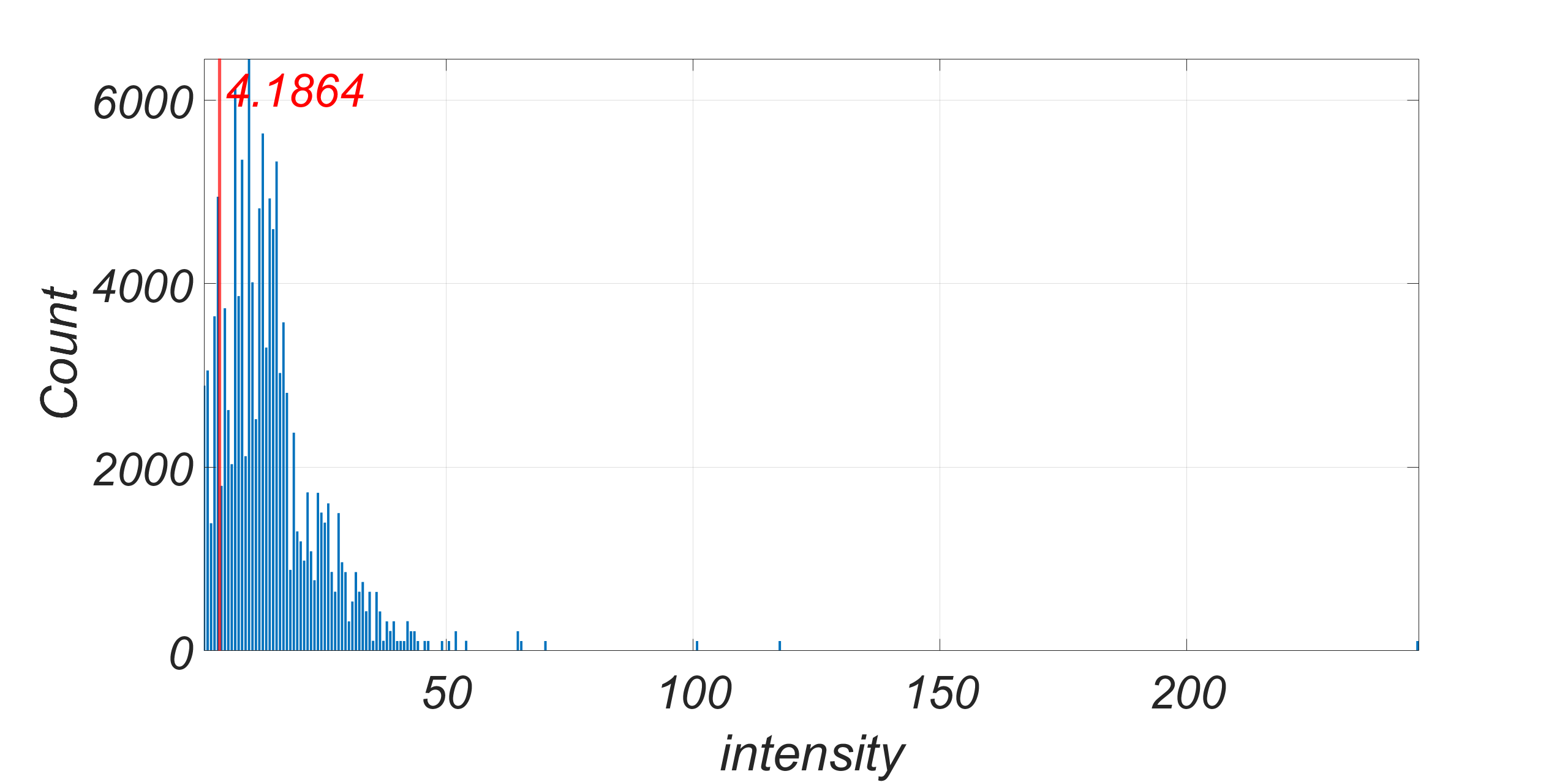}
        \caption{Histogram of~\gls{lidar} intensity for a).}
        \label{fig:fig1_filtered_weib}
    \end{subfigure}
     \begin{subfigure}[b]{0.315\textwidth}
        \centering
        \includegraphics[width=\columnwidth]{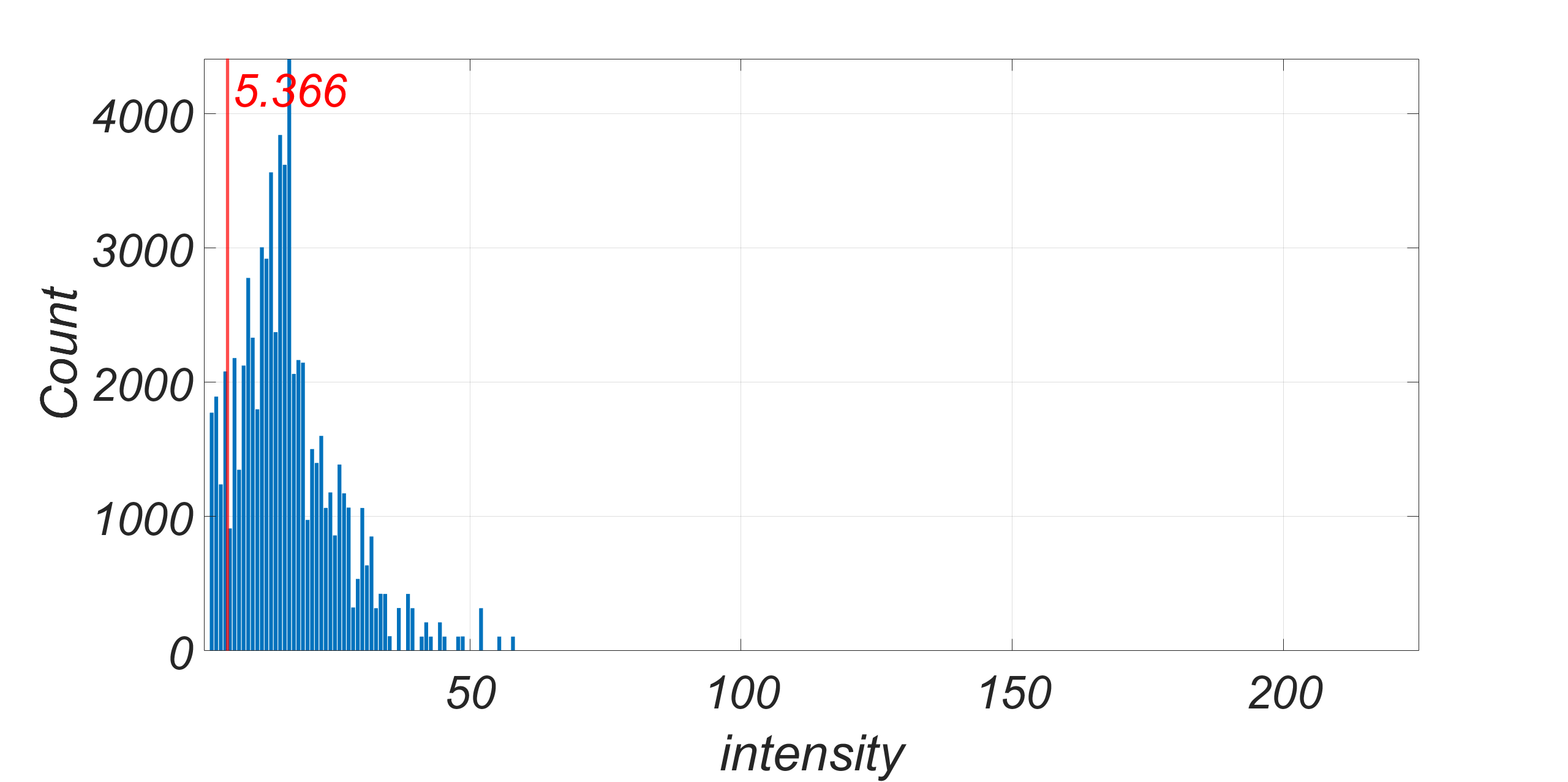}
        \caption{Histogram of~\gls{lidar} intensity for b).}
        \label{fig:fig1_raw_pcl_weib}
     \end{subfigure}
     \begin{subfigure}[b]{0.315\textwidth}
        \centering
        \includegraphics[width=\columnwidth]{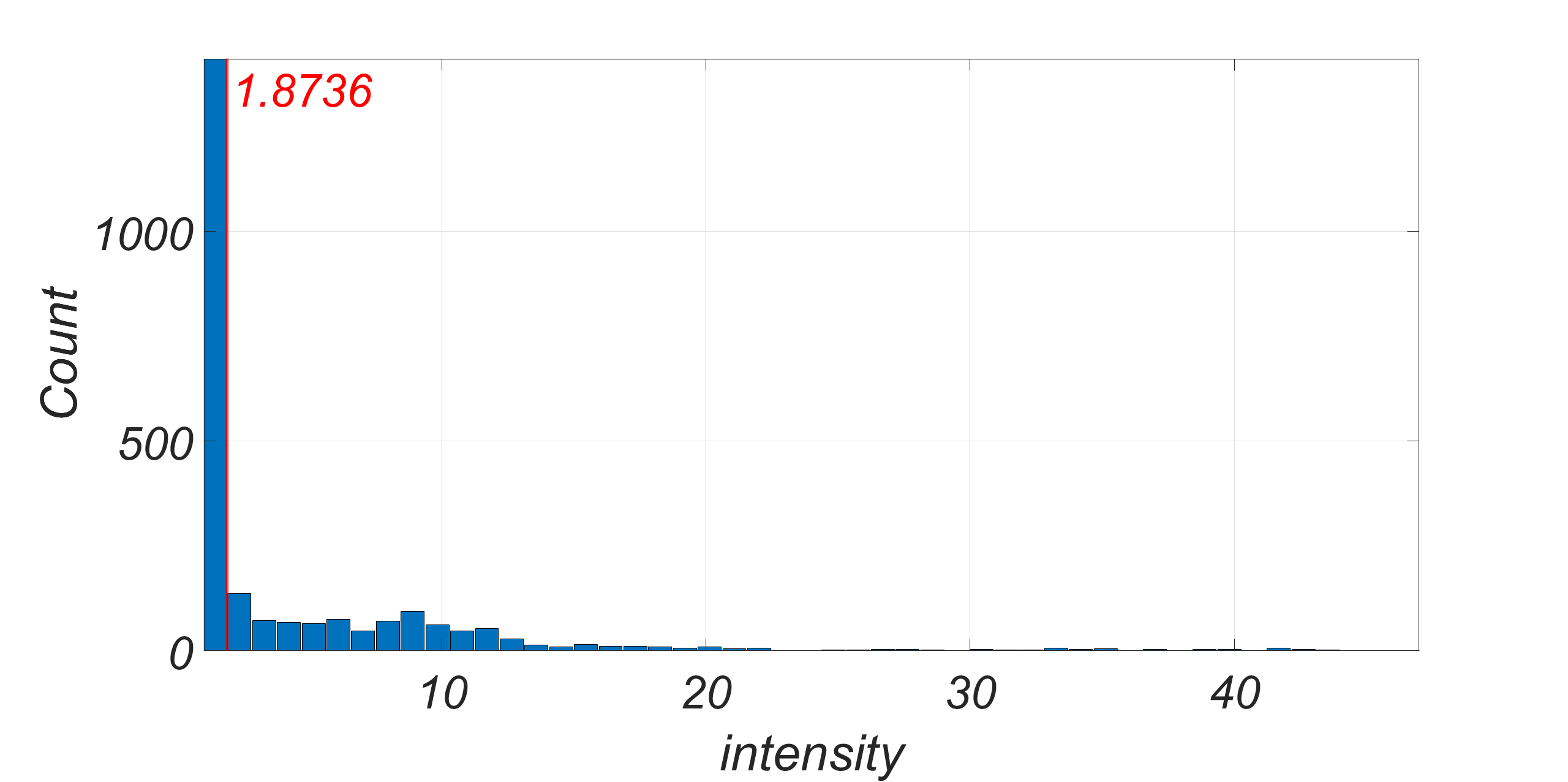}
        \caption{Histogram of~\gls{lidar} intensity for c).}
        \label{fig:fig1_dust_weib}
     \end{subfigure}

     \setlength{\abovecaptionskip}{4pt}
     \setlength{\belowcaptionskip}{-19pt}
     \caption{Test 1. Row 1: Generated \gls{pcl} map.~\textit{Left}, original~\gls{pcl},~\textit{middle}, filtered~\gls{pcl} and isolated dust,~\textit{right}. Row 2: Weibull distributions for the~\gls{pcl}s.}
     \label{fig:mapexample}
\end{figure*}

Classical methods such as~\gls{ror}~\cite{jia2016ror} and~\gls{sor}~\cite{carrilho2018sor} mostly rely on spatial information within~\gls{pcl}s to remove outliers based on the local density and distribution of points. This is feasible as points resulting from the aerosol particles have relatively low density compared with their neighboring clusters~\cite{prio2022dror}. However, these methods do not perform reliably in varying density~\gls{pcl}s, where the density of~\gls{pcl}s is proportional to the measured range. Therefore,~\gls{dror}~\cite{charron2018dror} and~\gls{dsor}~\cite{akhil2021} are proposed to improve the shortcomings of previous methods by dynamically adjusting the radius based on the range of any given points from the~\gls{lidar} to preserve more environmental features~\cite{prio2022dror}. Moreover, by utilizing inherent information such as intensity, aerosol particles can be further identified in~\gls{lidar} data~\cite{fritsche2018fusion} as they tend to have low intensity due to the absorption and refraction of most of the beam emitted by~\gls{lidar}~\cite{ali2022dust}. Thus, to further improve~\gls{lior}~\cite{park2020fast} and previous dynamic methods, a combination of the~\gls{dror} and~\gls{lior} is proposed~\cite{ali2022dust} to not only address the sparsity issues in the~\gls{lidar} data but also improve the $F1$-score when compared to the previous methods. Compared to~\gls{dror},~\gls{lidror} has lower time complexity due to the removal of points from~\gls{pcl} at long-range as well as fewer computational operations which resulted from the initial thresholding based on~\gls{lior}~\cite{park2020fast}.

Learning-based methods employ both traditional~\gls{ml}-based algorithms such as~\gls{knn}~\cite{duan2018weighted} and~\gls{dbscan}~\cite{zhao2018dbscan} as well as current deep-learning approaches that are based on various~\gls{nn} architectures~\cite{heinzler2020cnn} to perform either point- or voxel-wise classification~\cite{zhou2022fullreview}. Additional information such as remittance is also used to further improve the classification of such networks based on the correlation between material type and its reflectance~\cite{reymann2015improving}. Furthermore, both the spatial and temporal information is exploited to classify the outliers in~\gls{pcl} by utilizing motion-guided attention blocks~\cite{alexandru2021latticenet}. By leveraging the semantic segmentation networks that are well-established, a voxel-wise classifier is capable of segmentation of aerosol particles in the~\gls{pcl}s~\cite{alvari2023net}. This limits the scope of their usage onboard~\gls{mav}s, where due to the payload limitations, only a simple lightweight computing unit can be deployed.

Moreover, other approaches based on~\gls{ldmm}~\cite{osher2017low} and~\gls{mls}~\cite{guennebaud2008dynamic} are proposed that exploit the self-similarity of patches from~\gls{pcl}~\cite{zeng2020laplacian}. However, neither of these outlier removal methods can perform reliably in real-time~\cite{zeng2020laplacian}. Therefore, a more robust and computationally efficient framework is required for onboard deployment for autonomous navigation of heterogeneous-robotic systems.

In this paper, a novel modular agnostic filtration framework is proposed to dynamically remove points in~\gls{lidar}~\gls{pcl}s resulting from the presence of aerosol particles based on the combination of statistical outlier detection and smoothing filter. Furthermore, the proposed framework is directly coupled to the velocity of the platform and the density of the~\gls{pcl} while performing dynamic down-sampling to ensure low latency and computational complexity.

\section{Contributions} \label{contribution}
The main contributions of this study are as follows: (a) An online platform-agnostic modular filtration framework for~\gls{lidar} data based on both inherent and spatial information from~\gls{pcl} that relies on~\gls{doscor} approach, which runs solely on CPU. (b) Integration and coupling of the velocity of the robotic platform and time complexity of the algorithm for an adaptive activation of modules for close and long-range filtration. (c) Utilization of~\gls{sg} filter to perform outlier rejection via smoothing of the data in $1D$. (d) Evaluation of the proposed framework and its viability for field deployment and utilization in conjunction with collision avoidance method based on the~\gls{apf}~\cite{lindqvist2021compra}. (e) Experimental performance evaluation of the proposed approach in~\gls{sub-t} environments in the presence of aerosol particles.

\begin{figure*}[!t]
    \centering
    \includegraphics[width=1\textwidth, trim={2.5cm 19cm 4.5cm 4.6cm}, clip, page=1]{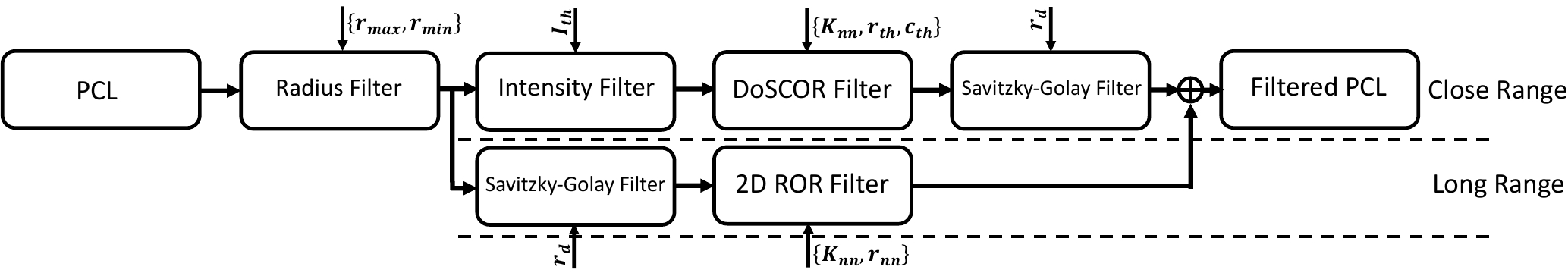}
    \setlength{\abovecaptionskip}{-7pt}
    \setlength{\belowcaptionskip}{-19pt}
    \caption{Overview of the~\gls{pcl} filtration framework.}
    \label{fig:framework}
\end{figure*}

The remainder of this article is structured as follows. Section~\ref{methodology} presents a detailed description of the implemented framework and discuss individual modules of the pipeline and their corresponding functionalities. Furthermore, experimental evaluation of the proposed pipeline and its effect on the obstacle avoidance algorithm is presented in Section~\ref{results}. Finally, we conclude this article by discussing the achieved results and future work in Section~\ref{conclusion}.

\section{Methodology} \label{methodology}
In this Section, the proposed modular outlier detection and removal method that is based on the dynamic clustering and~\gls{sg} is presented. The overall proposed pipeline with its subsequent processes is shown in Figure~\ref{fig:framework}. Element-wise pipeline descriptions and their computational impact are further analyzed and presented in this section.

\subsection{Environmental Noise Characterization}

As shown in Figure~\ref{fig:mapexample}, aerosol particles such as smoke and dust in~\gls{sub-t} environments make an evident impact on~\gls{slam} algorithms. The resulting maps of the environment based on the acquired~\gls{pcl}s in such conditions are cluttered and noisy and prevent optimal exploration and detection of survivors in~\gls{sar} missions. Figure~\ref{fig:sutb-dust} illustrates an example of such environments, where the proposed framework is assessed in the presence of aerosol particles.

As depicted in Figure~\ref{fig:dust-drone-concept}, the behaviors of the~\gls{lidar} scans in the presence of aerosol particles such as dust and smoke are directly related to the decomposition of aerosol particles, the~\gls{lidar} beam characteristics, and the material properties of the environment where the robot is located~\cite{phillips2017dust}. These behaviors can be summarized as no obstruction, full obstruction, partial obstruction, and full loss of~\gls{lidar} data~\cite{phillips2017dust}. In no obstruction cases, due to the low density of aerosol particles, the~\gls{lidar} can penetrate through the particles and detect obstacles. However, this is not feasible when the~\gls{lidar} beam is fully absorbed. Therefore, in the presence of high-density aerosol particles, it is feasible to falsely classify such cloud formations as either a part of the environment or obstacles. As such, the combined effect results in noisy and cluttered~\gls{pcl}s with possible occlusions that prevent optimal navigation of the~\gls{pde}s. Subsequently, other perception algorithms that rely on \gls{lidar} data such as~\gls{slam}, object detection, and collision avoidance would be directly affected, and the resulting interference diminishes the operational capabilities of any robotic platform~\cite{alvari2023net}.

\subsection{Framework Architecture}

The proposed framework solely requires unorganized unlabeled~\gls{pcl}s and their $xyz$-coordinate values as well as~\gls{lidar} intensity associated with these points. The scan measurements captured by the onboard 3D~\gls{lidar} are represented as a~\gls{pcl} in the~\gls{ccs} and are defined according to~\gls{ros} coordinate conventions as $x$ - forward, $y$ - left and $z$ - up.
The acquired~\gls{pcl}s tend to have noise induced by small particles that can be easily entrained with airflow from, for instance, a drone propulsion system as shown in Figure~\ref{fig:sutb-dust}. Thus, the main objective behind the proposed framework is the filtration of noise induced by these particles in three stages as shown in Figure~\ref{fig:framework}, where at first radius-based filtration in a~\gls{scs} is applied, next intensity-based filtration in~\gls{ccs} is utilized to remove low-intensity points from~\gls{pcl}, thereafter~\gls{sg} is applied to ranging measurements in~\gls{scs} to remove outliers and finally 2D~\gls{ror} filter in~\gls{ccf} is used to ensure that the spatial information of the small objects is preserved.

\subsection{\gls{lidar} Filtration}

The~\gls{lidar} filtration in this paper is implemented in Python within~\gls{ros} to enable its integration with other frameworks for online deployment. To ensure real-time performance with low computational complexity and deployment in all possible platforms, the input~\gls{pcl} is divided into close and long-range segments to not only prevent loss of spatial information but also create pseudo attention for regions that are vital for operational safety of the heterogeneous robotic system in~\gls{hsi} setting. As shown in Figure~\ref{fig:framework}, each module requires a set of parameters to operate. Thus, the values and selection criteria of these parameters are provided in Table~\ref{tab:parameters}.

\begin{table}[!b]
\setlength{\abovecaptionskip}{2pt}
\setlength{\belowcaptionskip}{-11pt}
\caption{Required parameters for the proposed framework.}\label{tab:parameters}
\centering
\scalebox{0.878}{ \centering 
\begin{tabular}{p{1.3cm}p{1.6cm}p{5.8cm}l|c|l}\toprule
        \textbf{Parameters}    &{\textbf{Initial Values}} &{\textbf{Hyper-parametric Conditions}}  \\ \hline \hline
        $r_{max}$       &{\unit[30]{m}}   &{$r_{max} = \{max\{d_{lon}\} | d_{lon} \in [10, 100]\}$}\\ \hline
        $r_{min}$       &{\unit[5]{m}}     &{$r_{min}= \{max\{\tau\} | \tau \in [2, 10]$\}}\\ \hline
        $I_{th}$        &{2}   &{$I_{th} = \{Q(p, \alpha, \gamma) | max\{p\} \in [0.1, 0.15] \}$}, Eq.~\ref{e:Eq_4b}\\ \hline
        $r_d$           &{\unit[\{4, 20\}]{m}}  &{$r_d = \{min\{d \} | d \in [r_{min}-1, r_{max}-10] \}$} \\ \hline
        $K_{nn}$        &{6}    &{$K_{nn} = \{min\{k\} | k \in [3, 6] \}$}\\ \hline
        $r_{th}$        &{\unit[0.45]{m}}   &{$r_{th} = \{ min\{r\} | r \in [0.2, 0.6]\}$}\\ \hline
        $c_{th}$        &{0.4}   &{$c_{th} = \{ min\{c\} | c \in [0.1, 0.5] \}$}\\ \hline 
        $r_{nn}$        &{\unit[0.15]{m}}     &{$r_{nn} = \{ min\{r_{nn}\} | r_{nn} \in [0.1, 0.16]\}$}\\ \hline \bottomrule
     \end{tabular}
     }
\end{table}
\subsubsection{Radius-based Filtration}

To achieve safe traversal in~\gls{pde}s,~\gls{rss} model~\cite{kim2021rss} is used to perform radius-based filtration. This allows overall performance improvement by minimizing the number of points required to be processed for navigation while addressing operational safety concerns. To utilize the~\gls{rss} model, coordinate system conversion from~\gls{ccs} to~\gls{scs} is performed. To obtain radial distance, $r$, inclination, $\theta$, and azimuth, $\phi$ based on $(x, y, z)$, axial radius, $\rho$, is given as $\rho = x^2 + y^2$. Thereafter, $r$ can be defined as $r = \sqrt{\rho + z^2}$ while $\theta$ and $\phi$ are calculated as $\theta = \tan^{-1}\left(\frac{\sqrt{\rho}}{z}\right)$ and $\phi = \tan^{-1}\left(\frac{y}{x}\right)$, respectively. The radius thresholds $r_{max}$ is selected such that it satisfies the following conditions:

\setlength{\abovedisplayskip}{-8pt}
\begin{align} \label{e:Eq_1}
  r_{max} = max \{d_{lon}\},
\end{align}

where $d_{lon}$ represents the longitudinal safe distances from the center of the robotic platform. The longitudinal safe distance is given by:

\begin{align} \label{e:Eq_2} \textstyle{
  d_{lon} = v_r \eta + \frac{1}{2} a_{accel} \eta^2 + \frac{(v_r + \eta a_{accel})^2}{2a_{min_{brake}}} - \frac{(v_f)^2}{2a_{max_{brake}}} ,
}\end{align}

where $a_{accel}$, $a_{min_{brake}}$ and $a_{max_{brake}}$ represent the maximum acceleration as well as minimum deacceleration of the robot and maximum deacceleration of the dynamic obstacle in the environment, respectively. Moreover, $v_r$ is the current velocity of the robot and $v_f$ represents the velocity of the dynamic obstacle. The estimated response time of the dynamic obstacle in a poorly illuminated environment is given by $\eta$. Additionally, the minimum radius threshold, $r_{min}$, is adaptively selected based on the time complexity of the algorithm and environmental complexity such that the filtered close-range~\gls{pcl} contains a maximum of $30k$ points. This is achieved by periodically sampling the number of points in close-range~\gls{pcl}. The sampling frequency of~\unit[1]{Hz} is selected for surveying close-range~\gls{pcl} as abrupt changes in the environment are mission critical and they directly affect the safety of the robot and rescue workers.


\subsubsection{Intensity-based Outlier Removal}

Based on the~\gls{lidar} intensity analysis in the absence and presence of dust and smoke~\cite{fritsche2018fusion, park2020fast}, it has been shown that the intensity of~\gls{lidar} data can be used to facilitate filtration of the noise in~\gls{pcl} due to the presence of aerosol particles. Furthermore, the intensity information can be utilized to distinguish and identify the material of various surfaces based on their reflectivity~\cite{fritsche2018fusion}. Due to the scattering and absorption of the~\gls{lidar} beam by the particles, the intensity value of aerosol particles in~\gls{pcl} is low and their distribution can be characterized based on Weibull~\gls{pdf}~\cite{Helmholz2020weibull}. The~\gls{pdf} of the general Weibull distribution is as follows:

\setlength{\abovedisplayskip}{-3pt}
\setlength{\belowdisplayskip}{5pt}
\begin{align} \label{e:Eq_3}
\resizebox{0.91\hsize}{!}{$%
  P(x, \alpha, \gamma, \mu) = \frac{\gamma}{\alpha} \left( \frac{x-\mu}{\alpha} \right)^{\gamma-1} exp\left(-\left( \frac{(x-\mu)}{\alpha} \right)^\gamma \right),$%
}
\end{align}

where $\gamma$, $\mu$, and $\alpha$ represent the shape, the location, and the scale parameter, respectively. Moreover, $x$ and $\alpha$ values are subjected to $x \ge \mu$ and $\alpha > 0$. Given $\mu = 0$, the general Weibull distribution equation can be expressed in its standard two-parameter form and is defined as:

\setlength{\abovedisplayskip}{-3pt}
\setlength{\belowdisplayskip}{10pt}
\begin{align} \label{e:Eq_4}
  P(x, \alpha, \gamma) = \frac{\gamma}{\alpha} \left( \frac{x}{\alpha} \right)^{\gamma-1} exp \left(-\left(\frac{x}{\alpha}\right)^\gamma \right).
\end{align}

As shown in Figure~\ref{fig:mapexample} and Table~\ref{tab:intensity_thres}, the intensity distribution of the noise generated in the~\gls{pcl} due to the presence of the aerosol particles has Weibull distribution. This behavior is observed independent of particle type as well as environmental factors~\cite{fritsche2018fusion}. Therefore, the intensity outlier rejection threshold, $I_{th}$, has an adaptive nature for a given data stream and is based on the Weibull~\gls{qf}~\cite{murthy2004weibull}, which is calculated using the following equation:

\setlength{\abovedisplayskip}{-7pt}
\setlength{\belowdisplayskip}{5pt}
\begin{align} \label{e:Eq_4b}
  Q(p, \alpha, \gamma) = \alpha \left( -ln\left( 1-p \right) \right)^{\frac{1}{\gamma}},
\end{align}

where $p \in [0, 1]$ represents the probability value such that the calculated $I_{th}$ has less than or equal probability value, $p$, as shown in Table~\ref{tab:parameters}. Moreover, this behavior is observed in row two of Figure~\ref{fig:mapexample}, in which after the noise removal the intensity threshold has changed.

\begin{figure}[!t]
    \centering
    \begin{tabular}{cc}
    \includegraphics[width=0.47\columnwidth, trim={0cm 0.1cm 0cm 0cm}, clip]{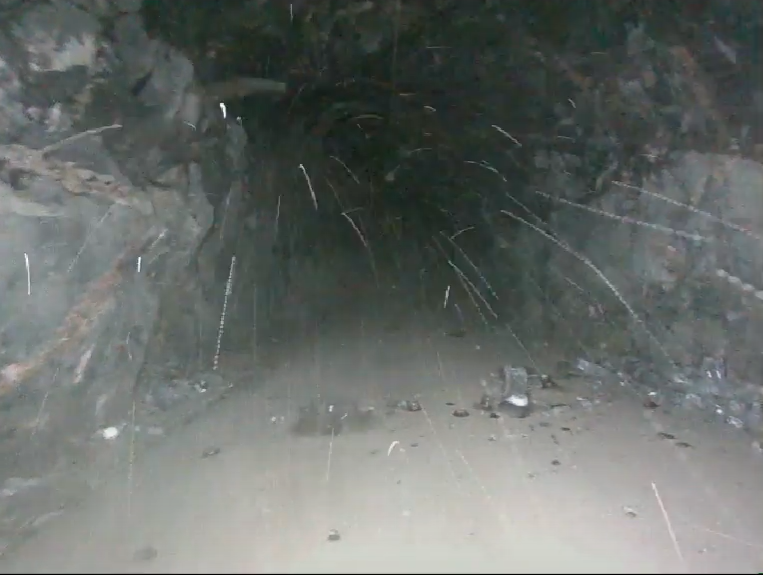}
    \includegraphics[width=0.47\columnwidth, trim={0cm 0.1cm 0cm 0cm}, clip]{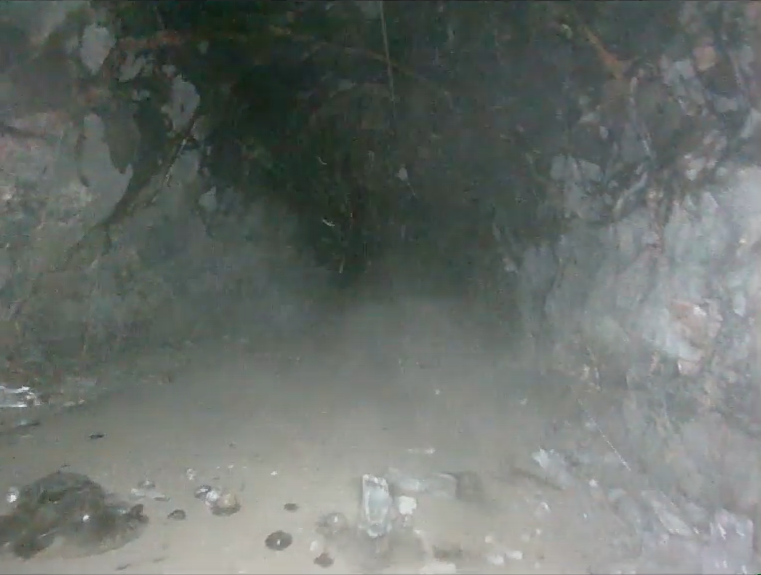}
    \end{tabular}
    \setlength{\abovecaptionskip}{-0.2pt}
    \setlength{\belowcaptionskip}{-19pt}
    \caption{Challenging~\gls{sub-t} environment for~\gls{lidar} sensors.}
    \label{fig:sutb-dust}
\end{figure}

\subsubsection{Dynamic onboard Statistical Cluster Outlier Removal}

The proposed~\gls{doscor} module is based on classical~\gls{ml} clustering method,~\gls{knn} and its combination with~\gls{sor} to enhance its capabilities while addressing the non-uniformity in~\gls{pcl}s. Kd-tree is used to structure the previously filtered 3D~\gls{lidar} data prior to performing an initial query ball-search within the radius of~\unit[0.05]{m} to detect the nearest neighbors of each point. This enables the analysis and characterization of individual points based on their spatial properties such as the distribution of neighboring points in~\gls{pcl} to identify and detect aerosol particles. Prior to the calculation of the distance vectors between the neighboring points and the resulting mean, $\mu$, and the standard deviation, $\sigma$, initial filtering is performed, where the points with $K_{nn} \le 6$ neighbors are removed from the~\gls{pcl}. To calculate the global static distant threshold, $s_{th}$, $\mu$ and $\sigma$ are calculated using the following equations:

\setlength{\belowdisplayskip}{1pt}
\setlength{\abovedisplayskip}{-8pt}
\begin{align} \label{e:Eq_5}
    \mu &= \frac{1}{n} \sum_{i=1}^{n} d_i, \\
    \sigma  &= \sqrt{\frac{1}{n-1} \sum_{i=1}^{n} \left( (d_i - \mu)^2 \right)},
\end{align}

where $n$ is the number of remaining neighbors for each point $P_i=(x,y,z)$ and $d_i$ is the Euclidean distance measured between point $P_i$ and its neighbors. The global distant threshold can be formulated as $s_{th} = \mu + (\sigma \cdot c_{th})$, where $c_{th}$ is a constant and its value is directly proportional to the density of aerosol particle present in the environment. Given the heterogeneous spatial distribution of points in~\gls{pcl} due to the 3D~\gls{lidar} spatial resolution and the limited number of beams, a dynamic threshold, $d_{th}$, is proposed that can be formulated as $d_{th} = (s_{th} \cdot d_i \cdot r_{th})$, where parameter $r_{th}$ is chosen based on the desired point rejection ratio such that points with high spatial variance are removed.

\subsubsection{\gls{sg} Smoothing and Outlier Removal}

Similar to commonly used approaches in robotics, where Kalman filter or~\gls{ml} is applied for regression,~\gls{sg} algorithm~\cite{savitzky1964smoothing} can be utilized to remove the outliers from~\gls{lidar} data via smoothing of~\gls{pcl} based on local least-square polynomial approximation in 1D. Denoting~\gls{lidar} range measurements, $r$, corrupted signal, $g(r)$, with additive noise, $\epsilon_r$, with zero mean and finite variance of $\sigma^2$, the $g(r)$ can be formulated by $g(r) = r + \epsilon_r$~\cite{sadeghi2020sgnew}. Based on this assumption, the~\gls{sg} polynomial fitting can be applied to the~\gls{lidar} range measurements to remove the noise resulting from the presence of the aerosol particles.

The~\gls{sg} smoothing method can be categorized as a kernel-based filtration due to its utilization of a symmetric sampling window of length, $w$, to compute and minimize the mean-squared error along the input range $g(r)$. The length of the sampling window is defined as $w= 2m +1$, where $w$ must be larger than the desired fitted polynomial degree $n$ to satisfy the minimum input constraint~\cite{savitzky1964smoothing}. Based on the least-square criterion, the summation of the squared differences between the observed range measurements, $r_i$, and the estimated polynomial, $p_i$, can be modeled as a cost function $\delta_m$, using the following equation:

\setlength{\abovedisplayskip}{-12pt}
\setlength{\belowdisplayskip}{4pt}
\begin{align} \label{e:Eq_6} 
    \delta_m =  \sum_{i=-m}^{m} (p_i - r_i)^2,
\end{align}

where $p_i = \sum_{k=0}^{n} b_{nk}i^{k}$. Moreover, the $k$th coefficient of the polynomial $p_i$ is denoted by $b_n$ and its value can be determined by differentiating $\delta_m$ with respect to $b_n$ and minimizing the resulting equation. This leads to the following equality equation:

\setlength{\abovedisplayskip}{-6pt}
\setlength{\belowdisplayskip}{4pt}
\begin{align} \label{e:Eq_7}
    \sum_{k=0}^{n} (\sum_{i=-m}^{m}  b_{nk}i^{k+j}) = \sum_{i=-m}^{m} ( i^{j} r_i),
\end{align}

where $j \in [0, n]$ is the index representing the equation number, given that there are $n+1$ equations. To calculate the coefficient vector $b$, the previous equality equation can be written in a matrix form as $ (\mathbf{A}^T \mathbf{A}) b_n=\mathbf{A}^T r_i$, where $b_n$ can be derived as $b_n = (\mathbf{A}^T \mathbf{A})^{-1} \mathbf{A}^T r_i$, given any sequence of range measurements, $r_i$, from~\gls{lidar}~\cite{sadeghi2020sgnew}. Furthermore, due to the non-uniformity of~\gls{pcl}, an incremental~\gls{sg} implementation is proposed to not only have varied radius threshold, $r_d$, for close and long-range~\gls{lidar} data but also varied polynomial degree, $n$, and window size, $w$, to prevent loss of spatial features by over-smoothing of the~\gls{pcl}. Overall, the larger window size and lower polynomial degree result in a higher degree of filtration~\cite{sadeghi2020sgnew}. Therefore, to achieve the desired noise removal, the optimal window size, $w_{opt}$, is calculated as follows:

\setlength{\abovedisplayskip}{-5pt}
\setlength{\belowdisplayskip}{7pt}
\begin{align} \label{e:Eq_8}
    w_{opt} = \left[ \frac{2(n+2)((2n+3)!)^2}{((n+1)!)^2)} \frac{\sigma^2}{\nu_n} \right] ^{2n+5},
\end{align}

where $\nu_n = \frac{1}{L} \sum_{l=1}^{L} (r^{(n+2)})^2$ and $L$ is the maximum number of sampling of the~\gls{lidar} data in a period $\tau$~\cite{sadeghi2020sgnew}.



\subsubsection{2D Radius Outlier Removal}

To minimize the computational cost associated with~\gls{ror} and preserve the environmental features~\cite{prio2022dror}, the 2D~\gls{ror} filter is implemented by projecting the previously filtered~\gls{pcl} into $XY$-plane thereby removing the $z$-axis spatial information from the~\gls{pcl} prior to construction of the Kd-tree for organizing the 2D~\gls{pcl}. Thereafter, a search query based on the radius, $r_{nn}$, is performed to detect the number of neighboring points prior to filtration. Based on the parameter, $K_{nn}$, which indicates the minimum number of acceptable neighbors for each point $P_d = (x,y)$ in the 2D~\gls{pcl}, the outlier removal is performed. Finally, the remaining points were merged with the close-range modules to produce the filtered~\gls{pcl} as shown in Figure~\ref{fig:framework}.

\section{Results} \label{results}
In this section, the evaluation of the proposed framework and its viability for deployment in real-world scenarios in~\gls{pde}s, specifically in~\gls{sub-t} environment located in Luleå, Sweden, is presented. To ensure the safety of both aerial and terrestrial robotic platforms during field experimentation, their traversal velocity is maintained at approximately~\unit[1.2]{m/s}. 





\subsection{System setup}

The sensor setup and the heterogeneous robotic system, which consists of an aerial and a terrestrial robot~\cite{lindqvist2022multimodality}, is utilized to evaluate the proposed filtration framework. Furthermore, both platforms were equipped with Intel NUC 10 BXNUC10I5FNKPA2.

\subsection{Evaluation of~\gls{slam} Methods in the~\gls{pde}}

\begin{figure}[!t]
     \centering
    \begin{subfigure}[b]{0.45\textwidth}
        \centering
        \includegraphics[width=\columnwidth, trim={0cm 0cm 0cm .92cm}, clip]{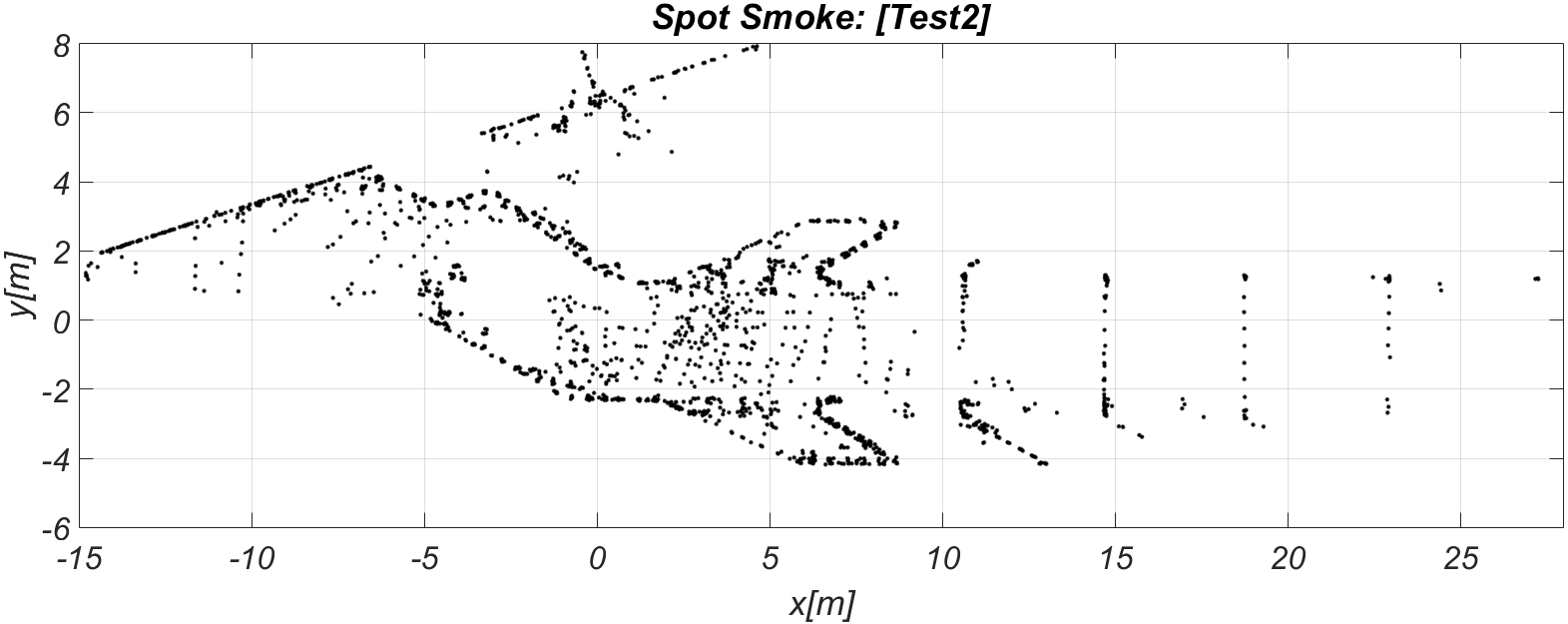}
        \setlength{\abovecaptionskip}{-10pt}
         \setlength{\belowcaptionskip}{2pt}
        \caption{Generated map based on~\gls{dlo} from raw~\gls{pcl} in the presence of smoke.}
        \label{fig:spot-filter-off2}
     \end{subfigure}
    \begin{subfigure}[b]{0.45\textwidth}
        \centering
        \includegraphics[width=\columnwidth, trim={0cm 0cm 0cm .92cm}, clip]{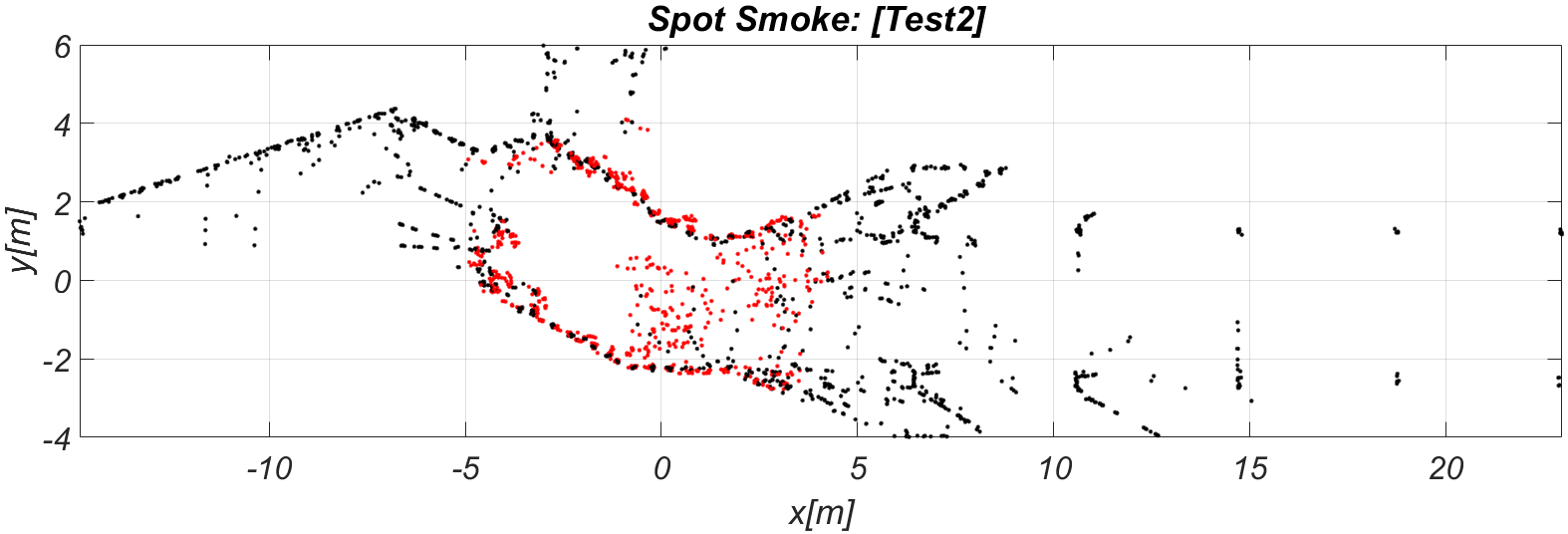}
        \setlength{\abovecaptionskip}{-10pt}
        \caption{Our framework on. Identified smoke in red, and the environment in black.}
        \label{fig:spot-filter-on2}
    \end{subfigure}
        \setlength{\belowcaptionskip}{-10pt}
        \caption{Test 2: \gls{dlo} map generation with and without our filtration framework.}
        \label{fig:smoke-filtering2}
\end{figure}

\begin{figure}[!t]
     \centering
     \begin{subfigure}[b]{0.45\textwidth}
        \centering
        \includegraphics[width=\columnwidth, trim={6cm 5.2cm 6cm 7cm}, clip]{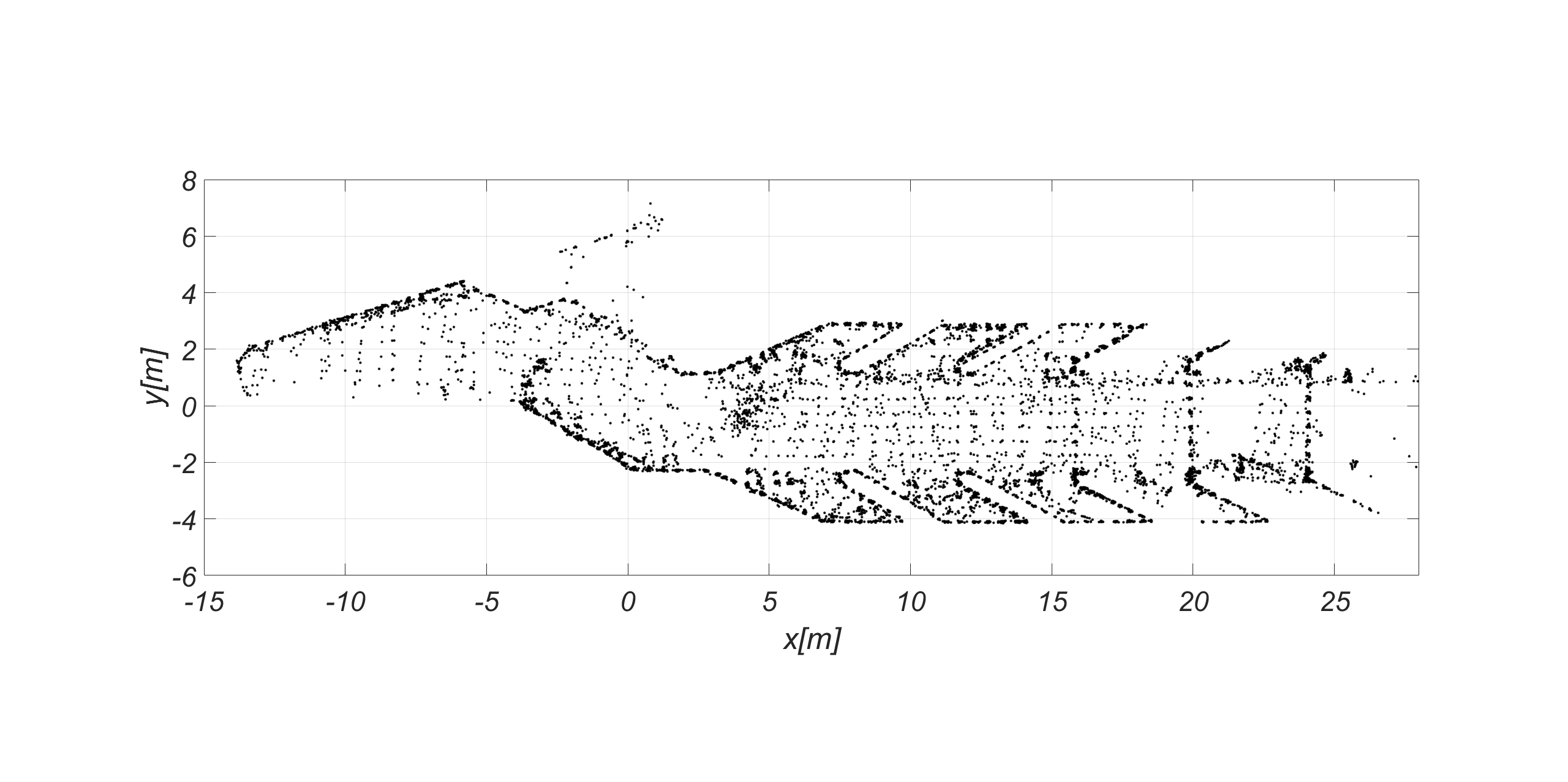}
        \setlength{\abovecaptionskip}{-10pt}
        \setlength{\belowcaptionskip}{2pt}
        \caption{Generated map based on~\gls{dlo} from raw~\gls{pcl} in the presence of smoke.}
        \label{fig:spot-filter-off3}
     \end{subfigure}
     \begin{subfigure}[b]{0.45\textwidth}
        \centering
        \includegraphics[width=\columnwidth, trim={6cm 7cm 6cm 9cm}, clip]{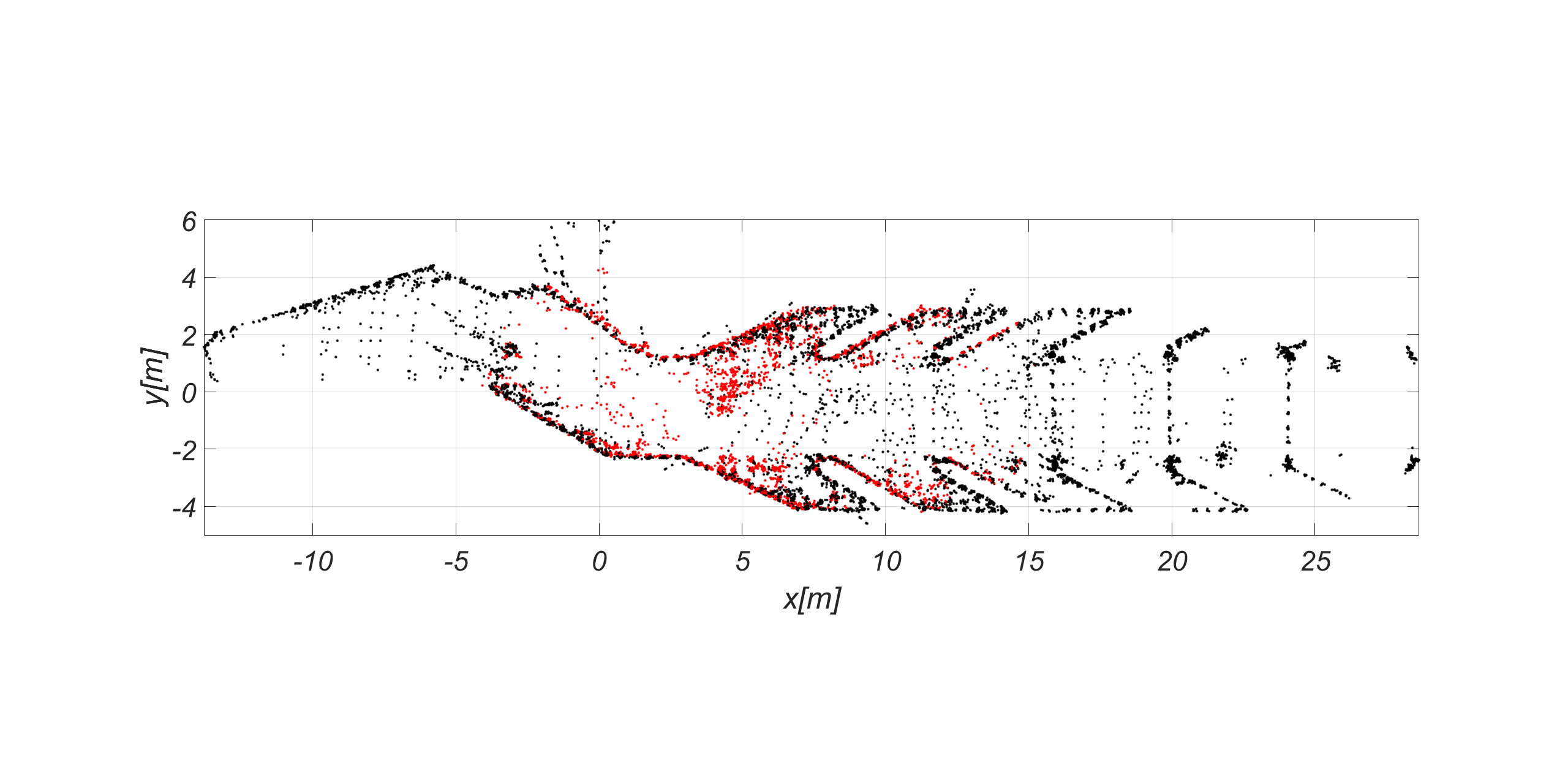}
        \setlength{\abovecaptionskip}{-10pt}
        \caption{Our framework on. Identified smoke in red, and the environment in black.}
        \label{fig:spot-filter-on3}
    \end{subfigure}
        \setlength{\belowcaptionskip}{-16pt}
        \caption{Test 3: \gls{dlo} map generation with and without our filtration framework.}
        \label{fig:smoke-filtering3}
\end{figure}

Current~\gls{soa}~\gls{slam} algorithms such as~\gls{dlo}~\cite{chenk2022} and~\gls{liosam}~\cite{shan2020lio} rely on onboard sensors to achieve localization and mapping of the environment by utilizing the fused~\gls{lidar} data in both temporal and spatial domain with~\gls{imu} data. To increase the fault tolerance, minimize the impact of noise from acquired~\gls{lidar} data, and generate fast and accurate key-frames, these algorithms utilize a combination of voxel-downsampling as well as an adaptive~\gls{icp} scheme based on~\gls{knn} and convex hull. However, these strategies are not sufficiently robust against aerosol particles in~\gls{pde}s as shown in Figure~\ref{fig:fig1_raw_pcl} with entrained dust, Figure~\ref{fig:spot-filter-off2} and Figure~\ref{fig:spot-filter-off3}, where the generated smoke is falsely identified as an obstacle and is included in the generated map of the~\gls{sub-t} environment.

Therefore, to assess the viability of the proposed framework and its impact on the generated map based on~\gls{dlo}, several comparative experiments as illustrated in Figure~\ref{fig:smoke-filtering2} and Figure~\ref{fig:smoke-filtering3} are performed in the presence of smoke, where the proposed framework was utilized as a pre-processing step for~\gls{pcl} prior to its utilization in~\gls{dlo}. Figure~\ref{fig:spot-filter-on2} and Figure~\ref{fig:spot-filter-on3} highlight the identified aerosol particles which were not included in the generated map when the proposed framework is utilized. By utilizing the proposed framework the entrained dust, Test 1, can be also detected as shown in Figure~\ref{fig:fig1_dust} and removed accordingly, as illustrated in Figure~\ref{fig:fig1_filtered}.


\begin{table}[!b]
\setlength{\abovecaptionskip}{2pt}
\setlength{\belowcaptionskip}{-11pt}
\caption{Dynamic intensity threshold based on the fitted Weibull distribution for outlier identification from various experiments.}\label{tab:intensity_thres}
\centering
\scalebox{0.966}{\centering
\begin{tabular}{p{1.3cm}p{3.1cm}p{2.36cm}p{0.7cm}l|c|c|l}\toprule 
        \textbf{Experiments}  &{\textbf{{Weibull~\gls{pdf} Parameters} [$\alpha, \gamma, \mu$]}}  &{\textbf{Intensity Threshold [$I_{th}$]}} &{\textbf{Classes}} \\ \hline \hline 
        $Test \; 1$       &{[${0.771938 ,\; 3.613051, \; 0.0}$]}  &{$1.873639$}&{51}\\ \hline
        $Test \; 2$       &{[${0.263573 ,\; 3.347880, \; 1.0}$]}  &{$5.276786$} &{28}\\ \hline
        $Test \; 3$       &{[${0.228001 ,\; 1.996240, \; 1.0}$]}  &{$4.614035$} &{57}\\\hline \bottomrule
     \end{tabular}
     }
\end{table}

\subsection{Evaluation of~\gls{apf} in the~\gls{pde}}

\begin{figure}[!t]
    \centering
    \includegraphics[width=\columnwidth]{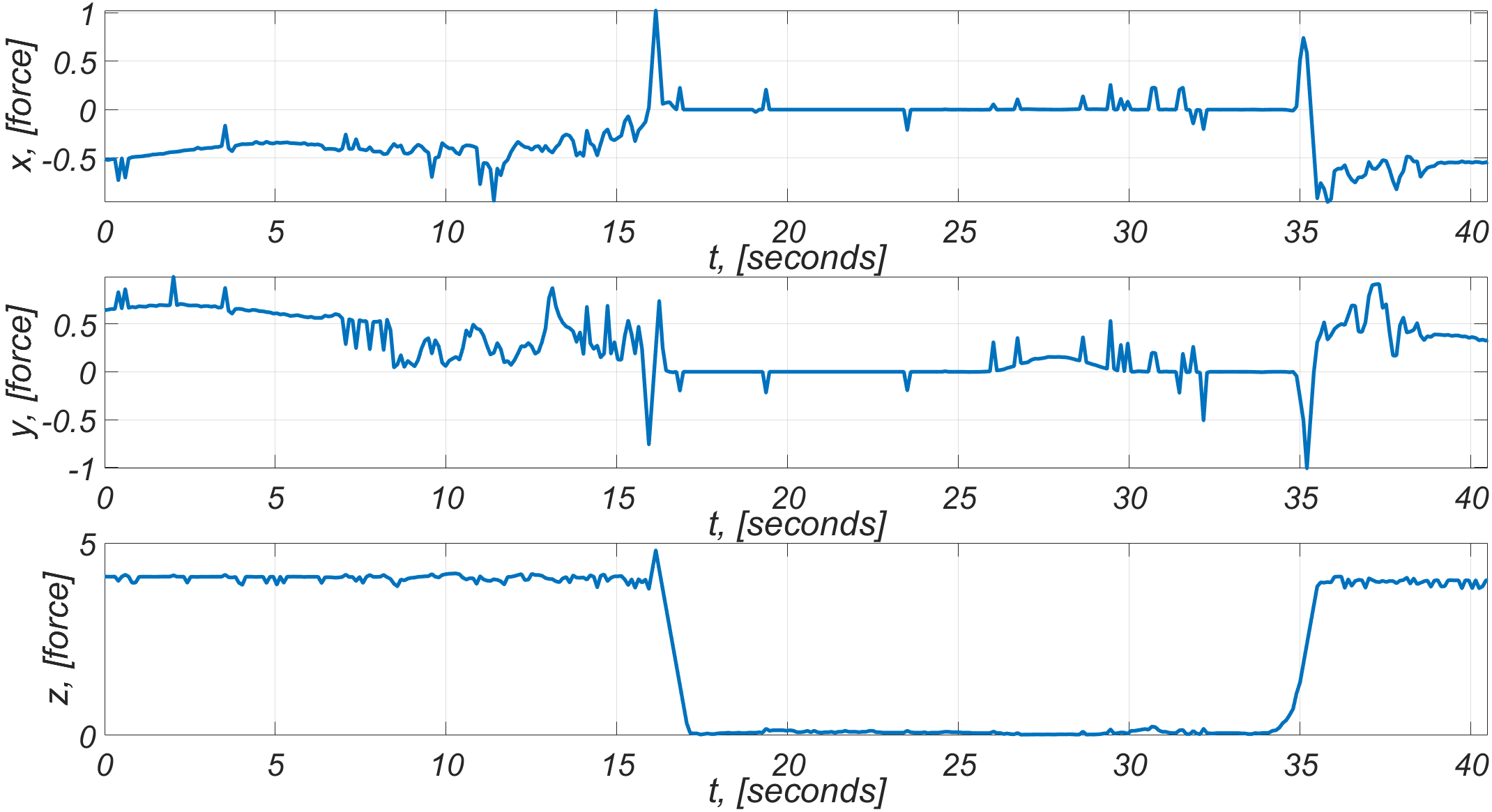}
    \setlength{\abovecaptionskip}{-8pt}
    \setlength{\belowcaptionskip}{-10pt}
    \caption{Potential field forces without the proposed framework.}
    \label{fig:apf-filter-off}
\end{figure}

\begin{figure}[!t]
    \centering
    \includegraphics[width=\columnwidth]{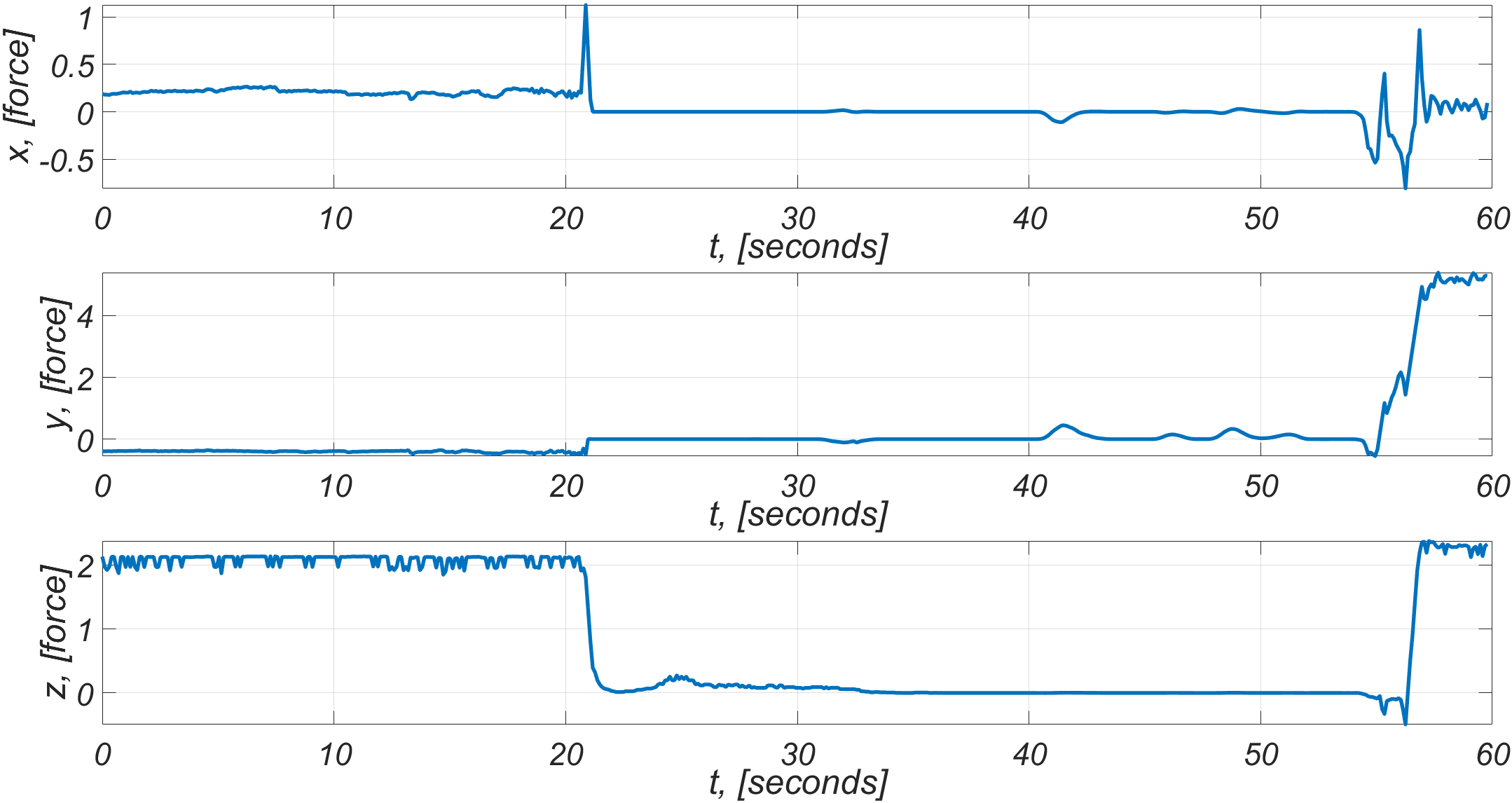}
    \setlength{\abovecaptionskip}{-8pt}
    \setlength{\belowcaptionskip}{-16pt}
    \caption{Potential field forces with the proposed framework.}
    \label{fig:apf-filter-on}
\end{figure}

By utilizing the~\gls{apf} method~\cite{lindqvist2021compra} in conjunction with the filtered and unprocessed~\gls{pcl} data stream, the impact of the proposed framework on obstacle detection and avoidance algorithms is studied. As shown in Figure~\ref{fig:apf-filter-off}, without the utilization of the proposed filtration scheme, the generated reactive forces have a higher variance when compared to their counterparts in Figure~\ref{fig:apf-filter-on}. Moreover, not only do the magnitude and direction of repulsive forces generated by~\gls{apf} differ in similar conditions in~\gls{pde} but also in several instances, the~\gls{apf} method could not detect the obstacles within the smoke thereby generating inaccurate forces, which can be observed in $y-$axis in Figure~\ref{fig:apf-filter-off}. It must be noted that such false classification will directly affect~\gls{sar} operations by limiting the scope of autonomous exploration and further hindering object detection algorithms to detect rescue workers and survivors in such harsh environments.



\section{Conclusions} \label{conclusion}
In this paper, an agnostic modular filtration framework for~\gls{lidar} data based on both the inherent and spatial information from~\gls{pcl} in harsh and unstructured~\gls{sub-t} environments with the presence of aerosol particles is presented. The proposed algorithm has demonstrated a solid performance in all experiments having an operational frequency between~\unit[10]{Hz} and~\unit[20]{Hz} for outlier detection and removal in a scalable pipeline for denser~\gls{pcl}s. Further performance gains can be realized by optimizing clustering methods, implementing dynamic non-uniform regional-based down-sampling of~\gls{pcl} and by implementing the current framework in low-level language like C++. The addition of spatial-temporal coupling can further improve outlier detection and removal by identifying static features thereby isolating noise and dynamic obstacles from the environment. Finally, other unsupervised clustering methods such as HDBSCAN and Local Outlier Factor (LOF) can be further evaluated and their performance, when compared to the current framework, investigated to achieve either better noise isolation or further time complexity improvement of the proposed framework.


\bibliography{IEEEabrv, main}
\end{document}